# 育野：
# 技术多样性与机器中的"野"
# Cultivated Wildness:
# Technodiversity and Wildness in Machines

张子豪
弗吉尼亚大学建筑学院构建环境博士候选人、景观系讲师

耿百利
弗吉尼亚大学建筑学院景观设计学系教授、系主任

ZHANG Zihao*
PhD Candidate in Constructed Environment and Lecturer in Landscape Architecture, School of Architecture, University of Virginia

Bradley CANTRELL
Professor and Chair, Department of Landscape Architecture, School of Architecture, University of Virginia

* Corresponding Author
Address: Campbell Hall, School of Architecture, 110 Bayly Dr, Charlottesville, VA 22903, USA
Email: zz3ub@virginia.edu

**摘要**

本文基于景观设计和人工智能的交叉领域探究了"育野"这一理念。当代景观实践应克服针对"荒野"概念可能存在的狭义理解，借鉴环境人文学、科学技术论、生态科学及景观设计学理念，探索用以培育新型野地的景观策略。通过剖析环境工程、计算机科学及景观设计案例，本文构想了利用智能机器营造"野地"的理论框架。该框架将机器视为具有能动性、能够参与合作生产的智能体，而非用来扩大人类对环境影响力的"数字基础设施"。近年来控制论技术发展迅速，传感网络、人工智能和网络物理系统等理念亦可为构建"育野"框架提供支持。该框架的核心是与"生物多样性"同等重要的"技术多样性"，以性能优化、追求效率为目标的技术发展单一愿景拘囿了人们的思想，剥夺了人们与环境相联系的其他可能。因此，育野的关键也在于认识机器的"野"。

**关键词**

人工智能；野性；景观设计；机器生态；技术多样性

**ABSTRACT**

This paper investigates the idea of cultivated wildness at the intersection of landscape design and artificial intelligence. The paper posits that contemporary landscape practices should overcome the potentially single understanding on wilderness, and instead explore landscape strategies to cultivate new forms of wild places via ideas and concerns in contemporary Environmental Humanities, Science and Technology Studies, Ecological Sciences, and Landscape Architecture. Drawing cases in environmental engineering, computer science, and landscape architecture research, this paper explores a framework to construct wild places with intelligent machines. In this framework, machines are not understood a layer of "digital infrastructure" that is used to extend localized human intelligence and agency. Rather machines are conceptualized as active agents who can participate in the intelligence of co-production. Recent developments in cybernetic technologies such as sensing networks, artificial intelligence, and cyberphysical systems can also contribute to establishing the framework. At the heart of this framework is "technodiversity," in parallel with biodiversity, since a singular vision on technological development driven by optimization and efficiency reinforces a monocultural approach that eliminates other possible relationships to construct with the environment. Thus, cultivated wildness is also about recognizing "wildness" in machines.

# 1 引言：从自然的社会建构到多元的"野"

除却环境问题层面针对荒野保护的科学探讨外，"荒野"（wilderness）概念一直颇具争议。在《荒野的困境》这一引发了众多讨论的评论文章中，环境历史学家威廉·克罗农从历史角度批判了荒野概念，并梳理了20世纪中叶以来美国在荒野保护工作中存在的问题。[1]收录了克罗农对现代环保主义辛辣批判的文集——《不同的立场：对自然的重构》——也已成为环境人文学及其他交叉学科的重要论著。[2]该文集还收录了其他学者的观点，包括景观理论家安妮·惠斯顿·斯本、景观地理学家肯尼斯·奥维格、女性主义理论家唐娜·哈拉维和文学评论家N·凯瑟琳·海尔斯，他们当中的很多人如今都已成为人文社会科学和设计学科（包括景观设计学）翘楚。从该文集中可以窥见学者在社会文化和历史语境下对"自然"和"荒野"等概念的质疑，这些"反叙事"的论述极大挑战了当时的主流认知。

上述学术探讨是"自然的社会建构"运动的重要组成[2]-[6]，该运动反映了20世纪下半叶至21世纪初各个学科对科学事实和真理的广泛反思[7]-[12]。此次思潮的主要论点是：自然和荒野等概念的形成离不开社会建构，它们承载着某些特定的社会文化价值观，但忽视了主流叙事之外的人类和非人类物种。[2][5][6][13]-[16]受此影响，当下逐渐形成了一种批判自然和荒野等概念的主要框架——通过情景化和地方性知识观来挖掘反叙事策略，向与人类及环境相关的主流观念发起挑战。[16][17]若不具备这种批判性视野，特定的荒野意象就会成为傀儡，任由各种环境实践中道德律令般的价值观肆意主导。某些特定文化的价值观不仅固化了对荒野的狭义认知，也加剧了长久以来的社会和政治冲突。例如近来有关美国阿拉斯加地区石油开采权的争论，再如工人阶级利益与环保主义者保护"原始自然"行动的冲突，又如原住民保卫家园和殖民者开拓"想象中的荒野"之间的长期纷争。[18]-[21]

当然，这些批判性思考绝不是要否定人们为环境保护做出的努力，而是要揭露那些可能削弱当代环境实践道德准则的价值观。不断发展的批判性观念正融入当代跨领域的环境正义运动中，积极推动设计师、环境工程师和环保主义人士反思其价值观并拓展对自然和荒野的认知。例如，美国山峦协会和荒野保护协会等环保组织已经拓宽对自然和荒野的定义，并致力于贯彻环境实践的包容性和公平性；近期

# 1 Introduction: From Social Construction of Nature to Plurality of Wildness

Outside the scientific environmentalist conservation and preservation discourse, "wilderness" has always been a contentious concept and a contested ground. In his (in)famous essay "The Trouble with Wilderness," environmental historian William Cronon issued a critique of the notion of wilderness from a historical lens and teased out the problematic aspects of America's wilderness preservation efforts since the mid-20th century.[1] Cronon's provocative critique of modern environmentalism is the leading essay in an important anthology in and outside environmental humanities—*Uncommon Ground: Toward Reinventing Nature*.[2] Other authors include landscape theorist Anne Whiston Spirn, landscape geographer Kenneth Olwig, feminist theorist Donna Haraway, and literary critic N. Katherine Halyes. Many of them later became influential and widely cited scholars across humanities, social sciences, and design disciplines, including Landscape Architecture. The anthology contributors interrogated the conception of nature and related concepts, including wilderness, by situating the terms within socio-cultural and historical contexts, revealing the counter-narratives that challenge mainstream beliefs.

This body of scholarship can be situated in a movement known as "social construction of nature,"[2]-[6] which mirrors the broad-based reflection on scientific facts and truth across disciplines in the second half of the 20th century and early 21st century.[7]-[12] The major argument formed wherein was that concepts such as nature and wilderness are at least partly constructed by the society and laden with specific socio-cultural values, overlooking people and nonhuman species that are outside the mainstream narratives.[2][5][6][13]-[16] This becomes the major analytical framework of today's critique on the conceptions of nature and related concepts including wilderness: unearthing counter-narratives with situated and localized knowledge to challenge the mainstream conceptualizations of human and the environment.[16][17] Without a critical lens, the image of a specific wild place could become a vessel into which the society pours all kinds of values that serve as moral imperatives for environmental practices. These culturally specific values reinforce a single vision of wild places, underpinning ongoing social and political conflicts, such as recent political debates on oil rights in Alaska, disputes between the working class and environmentalists' effort to protect "pristine nature," as well as long-term contentions rooted in indigenous people's home and colonialists' "wilderness imaginary."[18]-[21]

Of course, these critical reflections are never meant to undermine environmental efforts but only reveal the underpinning values in contemporary environmental practices that may undercut moral imperatives. These critical arguments keep developing and feed into contemporary environmental justice movements across fields, impelling designers, environmental engineers, conservationists, and preservationists to reflect their values and conceptions about nature and wilderness. For example, many environmental groups, such as Sierra Club and the Wilderness Society, have been addressing their ongoing inclusion and equity efforts in preservation and conservation practices; Also, recent land acknowledgment

北美各种机构开展的领土权属确认运动也肯定了原住居民长久以来的土地管理者身份。

随着自然和荒野概念的拓展，不同领域的学者开始以多元化的视角来理解和定义"野地"。在人类世的背景下，生态学家和生物学家开始将研究重心转移到"新形式的野"（new wilds）和人类世生态系统（novel ecosystems）上。[22]-[25] 近年来提出的环境策略也已不再强调维持或恢复历史生态格局，而鼓励通过促进生态系统过程、非人类物种和智能机器的自主性来实现"再野化"。[26]-[28] 人文和科学技术论等领域的学者也不再批判自然和荒野，转而以"非人类能动性"（nonhuman agency）为概念框架来理解人类与非人类物种及机器之间的合作生产和共同进化过程，探究三者间不同的关联方式。[29]-[34] 与此同时，景观设计师和规划师也不断吸纳新型生态系统、多物种相互作用、机器智能和非人类能动性等新兴思想，以期畅想更具开放型的设计项目。[35]-[39]

在承认荒野概念的核心在于自然的自主性（无或较少人工干预）的前提下，我们也需要认识到，机器在与其他自然过程和非人类物种的不断融合中，能够甚至已经发展出新颖有趣的荒野概念，并产生了重要的社会和生态影响。本文提出"育野"（cultivated wildness）概念，期望通过传感网络、人工智能和网络物理系统等前沿数字技术，探究有助于培育新型野地的全新景观策略，为有关荒野的讨论提供新思路。

通过剖析环境工程、计算机科学及景观设计和研究中的案例，本文认为，智能机器并非人类用以扩大自身（构想的）能动性或统治非人类物种的控制机制；它们作为多尺度的参与者，深深嵌入各种环境过程中，共同创造出超越人类认知范围的各种形式的"野"。换言之，景观设计师如今所需考虑的"环境"已经与过去大相径庭——它不再是可供设计师在其上尽情畅想各种景观动态变化的背景板或白纸，而是一种遍布各种智能的"反馈环境"。因此，"再野化"不再仅仅意味着利用有限的人类智能来恢复历史生态格局，或根据特定社会团体的标准来固化"荒野"意象。我们可以通过提升生态系统、非人类动植物物种和智能机器的自主性，精心培育新形式的野。当这些智能体发展出自己的路线、轨迹和变化过程后，便形成了真正开放、未知的荒野。

endeavors across North American institutions have recognized indigenous peoples as the longest-serving stewards of the land.

With the thickening of the conceptions of nature and wilderness, scholars across fields have begun to embrace a plurality in interpreting and conceptualizing wild places. Some ecologists and biologists have shifted their attention to "new wilds" and novel ecosystems in the face of Anthropocene.[22]-[25] Recent environmental strategies have surpassed maintaining or recovering historical ecological patterns and instead promote the autonomy of ecosystem processes, nonhuman species, and intelligent machines as strategies for "rewilding."[26]-[28] Scholars in Humanities and Science and Technology Studies (STS) have also turned their attention away from critiquing "nature" and "wilderness," and deployed "nonhuman agency" as a conceptual frame to understand the co-production and co-evolution between human and nonhuman species and machines, speculating different forms of associated relationships as well.[29]-[34] Landscape architects and planners also incorporate emerging ideas about novel ecosystems, multispecies interactions, machine intelligence, and nonhuman agency to envision a greater openness in their projects.[35]-[39]

Notwithstanding the essence of the wilderness concept as nature is "on its own" with little or no human guidance, there also is a sufficient need to understand how a convergence of machines with natural processes and nonhuman species can develop—and has already developed—new and interesting concepts of wildness with important socio-ecological ramifications. This paper contributes to this broad discourse about wilderness by proposing the idea of "cultivated wildness." Focusing on recent developments in cybernetics technologies such as sensing networks, artificial intelligence (AI), and cyberphysical systems, this paper attempts to demonstrate that such technologies can provide new opportunities for landscape strategies to cultivate new forms of wild places.

With examples drawn from environmental engineering, computer science, and landscape design and research, this paper argues that intelligent machines are not a layer of control mechanism through which humans extend imagined agency and expand an illusory control regime. Instead, intelligent machines should be conceived as multi-scalar actors that are deeply embedded in all kinds of environmental processes and give rise to all kinds of wild conditions outside human comprehension. In other words, today the concept "environment" is fundamentally different from what landscape designers considered and conceived in the past—Environment no longer stands only for a passive background and a tabula rasa on which designers entertain landscape dynamics. Rather, landscape architects have to confront a "cybernetic environment" with different forms of distributed intelligence. This paper argues that "rewilding" is no longer only about using localized human intelligence to restore historical ecological patterns and reinforce an image of wild place based on certain social groups' standards. Instead, "new wild" can be carefully cultivated by promoting the autonomy of ecological systems, nonhuman animals and plant species, and intelligent machines. In this way, by allowing the agents to take their courses, trajectories, and processes of becoming, wild places can be truly open-ended and indeterminate.

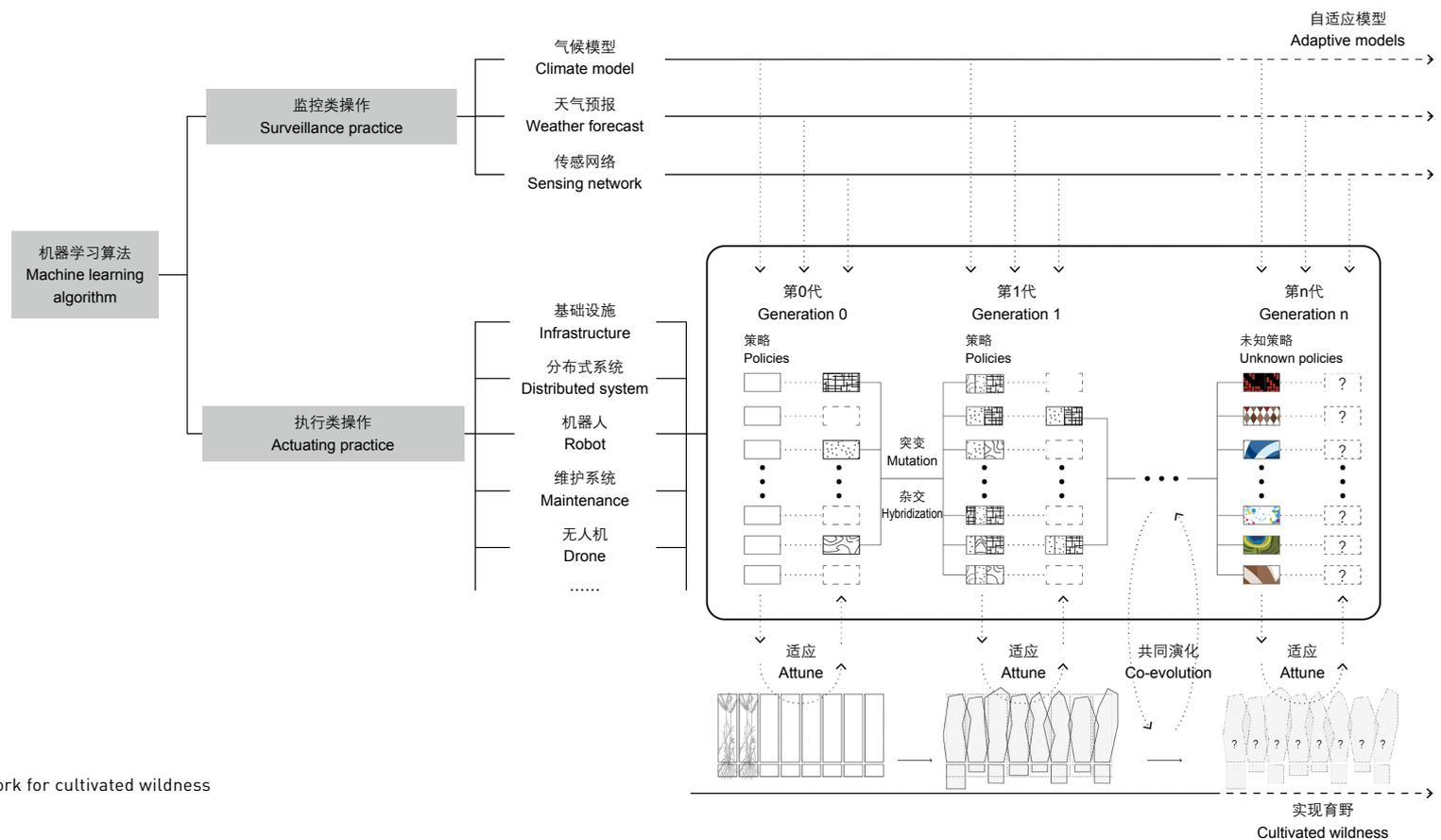

1. "育野"概念框架
1. Conceptual framework for cultivated wildness

本文提出一种"育野"概念框架（图1）：机器学习算法分布在不同的信息物理系统中——从环境监测系统（如气候模型、天气预报系统、传感网络等）到各种执行系统（如信息物理基础设施、机械装置、分布式机器人和无人机等）。这些智能体尺度各异，机械装置等小型执行器能够直接与植物相互作用，而加装了自动装置的水闸和溢洪道等大型执行器则可以通过改变土壤条件和水文格局来影响植物生境。机器学习智能体可通过持续学习和遗传算法来尝试各种不同策略，并通过迁移学习来彼此交换经验。机器学习技术还可以帮助建立更多的气候和天气适应性模型，这些模型的预报结果反过来又可用来训练分布式机器学习智能体。这些松散耦合的模型和智能体可以根据环境监测和过往经验来调整其策略；甚至随着时间的推移，提出超越人类理解范围的策略——即实现育野。通过开发与不同尺度的模型和系统相交互的界面，景观设计师可参与到这种新环境中。[40]此时，机器与其他非人类主体之间应当"相互依存"，而非"单向依赖"。在育野框架中，机器不仅是自动化工具，更是环境的共同创造者。正如

This paper maps out a conceptual framework to approach cultivated wildness (Fig. 1). Machine learning (ML) algorithms are distributed in cyberphysical systems from environmental surveillance practices such as climate models, weather forecasting systems, sensing networks to all kinds of actuating systems, including cyberphysical infrastructures, robotic armatures, distributed robots, and drones. These agents are multi-scalar, and some actuators (e.g. robotic armatures) can directly interact with the plants while the others (e.g. retrofitted flood gates and spillways) can impact plant habitats by modifying soil conditions and hydrological patterns. ML agents test out strategies through a framework based on continuous learning, genetic algorithms, and transfer learning. ML techniques also help build more adaptive climate and weather models, and the prediction results can feed into the training of the distributed ML agents. These loosely coupled models and agents could adjust their strategies, and, over time, would start developing unexpected strategies beyond human comprehension—cultivated wildness. Landscape architects can participate into it by developing different interfaces with these multi-scalar models and systems.[40] The relationship between machines and other nonhuman agents should be considered with the notion of interdependency rather than codependency. In this framework, machines are not tools for automation but have always been involved in and will play a preeminent role in the co-production of the environment. Eventually, constructing wild places with intelligent machines is about expanding landscape architects' understanding of the environment itself, just like

"阿尔法狗"拓展了棋手对围棋的理解一样，利用智能机器创造荒野的最终目的是加深人们对环境本身的理解。机器可以揭示和构建人类以往无法识别和诠释的环境格局。唯有如此，"培育的野"才能孕育人类当前无法想象的可能和未来。

提出上述育野框架的目的并非提供制造强效机器的技术蓝图，而是促进景观设计师理解和认识环境的复杂性，意识到除人类智能外还存在其他形式的智能。该框架源于上述生态学、景观设计学、环境人文学和科学技术论等学科中的技术和理论探索，要求设计师在本体论和认识论上进行根本性转变，将所谓的"环境"理解为不同框架、意图和目标函数互相作用后的混合体。不同的智能体——不论是人类或非人类，不论存活与否，不论是技术上的或生态的，不论是生物或非生物——都能相互调整、增强、抑制、利用与抵消。由此产生的环境融合了不同的视角，超越了每个个体的原始意图，无数的智能体精心培育着环境中的"野"。当代景观设计要不断拓展人们的审美体验范畴。[41][42]相应的景观策略也不应局限于保护和恢复层面，而需在如今与非人类领域如此疏远的文化中，孕育意料之外的机遇。因此，育野也意味着为当代城市居民培育一种新的审美观念，鼓励人们拥抱城市中的"野"并与之共生。

## 2 机器智能与构建的"野"

基于近年控制论技术的发展，我们可从三个方向探索上述育野框架：1）深度强化学习和机器策略；2）响应式景观和分布式智能；3）植物—机器交互。需要注意的是，育野的具体操作不能单独建立在以上任意一个方向上。诚然，这类育野实践的可行性仍待商榷，但下文示例从多视角对框架进行剖析，共同描绘了机器在景观策略中发挥重要作用的未来蓝图。

### 2.1 深度强化学习和机器策略

在过去几年中，人工智能研究突破不断，在深度强化学习（DRL）方面更是见证了众多变革，驱动了有关智能的更深层次的探索。就DRL而言，智能体可以观察环境状态并采取行动：如果某些行动被证明是有效的，那么智能体将得到奖励。近年来，科学家已经训练出了多个DRL智能体，其中包括"阿尔法狗"系列，它们精通围棋并击败了最高水平的人类棋手。[43]"阿尔法狗"的训练过程采用自对弈——除基本规则外，人工智能系统未曾学习任何人类棋手的对局方式；类似的应用也可见于对人工智能游戏选手AlphaStar的训练中。[44]这些DRL智能体超越了人类水平，探索出了人类未曾探知的全新策略。在某种程度上，它们基于机器对游戏的理解，提出具有"机器风格"的策略，从而扩大了人们对游戏本身的理解。

景观设计师和生态学家亦探究过机器学习在环境管理中的应用潜力。例如，在某项思维实验中，研究人员构想了一种基于DRL的人工智能系统——"荒野创造者"，该系统可自行制定策略并创建超越人类认知的野性空间。[29]实验挑战了人们对"荒野"的传统认知，为探

AlphaGo that has expanded players' understanding of the Go game. Machines may reveal and construct patterns in the environment that humans previously cannot recognize and translate. Only in this way cultivated wildness can become a reserve for possibilities and a future that we cannot think of now.

This framework is never a technical blueprint for building powerful machines, but a roadmap with which landscape architects can start to understand and appreciate the environment's complexity, recognizing different forms of intelligence other than human. It is deeply rooted in the aforementioned technical and theoretical explorations in Ecology, Landscape Architecture, Environmental Humanities, and STS. It challenges designers to take on an ontological and epistemological shift in understanding the environment as a mash-up of different frameworks, goals, and objective functions. Different agents—human or nonhuman, living or nonliving, technical or biological, biotic or abiotic—modify, amplify, depress, utilize, and cancel each other. Thus the resulting environment is always already a mash-up of different perspectives beyond every agent's original intention. Agents are carefully cultivating a sense of wildness in the environment. Contemporary landscape design should expand people's aesthetic categories.[41][42] Landscape strategies are no longer limited to preservation, conservation, and restoration but about promoting unexpected encounters in contemporary culture that is so alienated from nonhuman realms. Cultivated wildness entails cultivating a new sense of aesthetic among modern urban dwellers to embrace and learn to live with the wild in the city.

## 2 Machine Intelligence and Constructed Wildness

There are three venues to explore the framework of cultivated wildness regarding the recent development in cybernetic technologies: 1) deep reinforcement learning and machine strategies, 2) responsive landscapes and distributed intelligence, and 3) plant–machine interactions. It should be noted that none of these examples solely illustrate the kind of operation that can create wild places—In fact, it is doubtful that such a practice exists. However, examples in the following sections represent different aspects of the framework of constructed wildness, and together, they render a possible future where machines play an important part in the developing of landscape strategies.

### 2.1 Deep Reinforcement Learning and Machine Strategies

The past few years have seen breakthroughs in AI research, especially on deep reinforcement learning (DRL) which provides transformative cases for scholars to ask more profound questions about intelligence. In DRL, an agent can observe the state of the environment and then act: If the act turned out to be effective, then the agent would get rewarded. Over the past few years, scientists have trained many DLR agents, including the AlphaGo series that have mastered the Go game and beaten the best human Go players.[43] The training process used the self-play technique, which means that the AI system has played with itself based on basic

究机器在构建野性空间过程中的作用提供了新思路。在某种程度上，这些荒野的"野"正在于人类无法用既有模型来理解人类世生态系统。

上述思维实验在当年看起来或许离经叛道，但如今相关研究已相继展开。例如，美国弗吉尼亚大学Link实验室的科学家们即探索了训练DRL智能体来管理雨洪的方法。[45][46]实验根据弗吉尼亚州诺福克市的雨洪系统构建了一处模拟环境，系统中的两片汇水区分别与独立的蓄水池相连；蓄水池的出口处安装有可控阀门，可以灵活调节启闭程度（图2）。智能体通过调节两个阀门来控制蓄水池的水位，一方面可防止出现洪涝，另一方面可维持蓄水池的理想水位。

在每一个模拟周期中，DRL智能体都能够获取系统的最新状态，包括不同节点的水深及洪涝风险、未来24小时的降水预测和潮汐预报。根据这些数据，DRL智能体可通过相应策略调节阀门。模拟系统会依据不同的策略进行更新，如果策略有效，则智能体会得到奖励。该DRL智能体的奖励机制取决于它能否有效防止因潮汐和降雨产生的洪涝并维持理想水位。在整个实验过程中，科学家们利用真实的潮汐和降雨数据模拟出一系列雨洪事件来训练和测试DRL智能体。

DRL智能体不仅可以识别实时数据的细微变化，还能够实现实时控制，这是人类能力远达不到的。最重要的是，DRL智能体能够提出一些意想不到的策略：雨后蓄水池充盈，智能体会连续地打开或关闭阀门缓慢排水，以维持目标水位，此时下游或出现小型洪涝，但洪水总量仍在可控范围内。智能体的这一行为提醒着设计师，城市水系管理往往只能优化个别变量，而智能体则能另辟蹊径，利用洪水的频率而非水量控制来管理雨洪，提出了基于信息物理系统的全新策略。由于从未接触过任何人类已掌握的雨洪系统知识，DRL智能体的学习

rules, without learning from any human knowledge. Similar applications can be found in video games, such as AlphaStar.[44] These DRL agents have not only outperformed humans but also came up with novel strategies that human players had never thought of. In a way, they developed a machine understanding of the games and came up with strategies in a "machine mindset," expanding human understanding of the games.

Landscape architects and ecologists have also speculated on the implications of ML in environmental management. In a thought experiment, the researchers imagined a DRL-based AI system called "wildness creator" that can devise its own strategies and create wild places beyond human comprehension.[29] This thought experiment questions the notion of wilderness and provides an alternative way to conceptualize machines' role in constructing wild places. The places constructed by the machines are wild in the sense that they are novel ecologies for which humans do not have a model to comprehend.

This thought experiment may sound outlandish, but there are, in fact, already related research efforts. Scientists from the Link Lab at the University of Virginia have explored training DRL agents to control stormwater systems.[45][46] The scientists constructed a simulated stormwater system inspired by a system located in Norfolk, Virginia, USA. It consists of two sub-catchments connected to two separate retention ponds. At the outlet of each pond is a controllable valve that can be closed or open to any degree (Fig. 2). The agent can control the two valves to adjust the water volume in the ponds. The goal is to, on the one hand, prevent overall flooding in the system, and, on the other hand, maintain a desired water level in the retention ponds.

In each step, the DRL agent can observe the current state of the system, including the water depths and flooding rate on each nodes, as well as rainfall and tide forecast in next 24 hours. Based on the data, the DRL agent can come up with a control policy to adjust valves. The policy would update the system, and the agent will be rewarded if the action is desirable. This DRL agent's reward function is defined by how well the agent could prevent overall flooding (both from tide and rainfall) and maintain certain water depths in the ponds. Scientists used real-world rainfall and tidal data to produce a range of flood events to train and test the DRL agent.

The agent can not only respond to the subtle changes of the real-time data, but also achieved a level of real-time control beyond human capacity. Most importantly, the agent has devised some unexpected strategies. When the pond is filled after a rain event, the agent would open / close the valves repeatedly and slightly discharge water to maintain the target water level. Though this would cause minor floods downstream, the agent can keep the overall volume under a threshold. This reminds designers that urban water systems are often optimized with a few parameters, and the agent's unexpected behaviors—for example, exploiting the frequency dimension of floods—offer a new range of potential strategies based on frequency rather than volume to manage urban stormwater with cyberphysical systems. Since no human knowledge of the stormwater system was given to the DRL agent, the machine

2. Link实验室的DRL实验示意图（改绘自参考文献[46]）
2. Schematic diagram for Link Lab's DRL experiment (Adaptation source: Ref. [46])

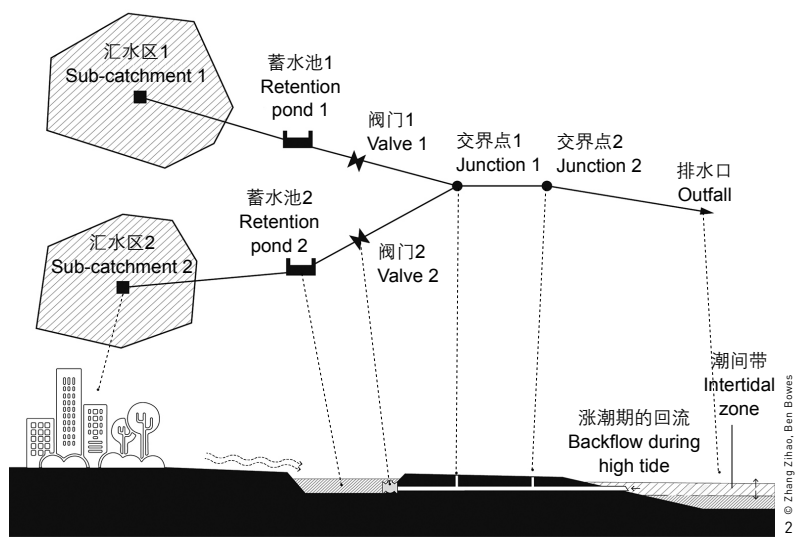

started from scratch and developed its own understanding of the system and came up with a set of strategies on its own.

In this example, the agent is trained to prevent floods. However, by understanding the importance of flood events to ecosystems, landscape architects can imagine a DRL agent trained to strategically promote flood events in certain areas in a city to create novel and emergent ecologies. Moreover, the DRL agent can actively maintain water depths in the retention ponds, which provide ideal conditions for wetland species.

More recently, scientists are looking to build more complex objective functions with multiple competing goals, such as water quality. Preliminary results suggest that the DRL framework can quickly build a diverse school of agents attending to different objectives (e.g. controlling flooding, managing pollutants), and there are many hybrid ones in the middle with complex behaviors. With this example, one can imagine multiple urban drainage systems managed by different agents with slightly different priorities. The feedback between these agents will give rise to complex phenomena beyond human imagination now. The agents' strategies would result in new hydrological patterns in the system; new ecological types would emerge through the DRL agents' careful cultivation. The result is a cultivated urban wild that is highly maintained by intelligent agents whose actions are sometimes beyond human comprehension.

## 2.2 Responsive Landscapes and Distributed Intelligence

Designers tend to ignore how deeply intelligent machines are already distributed in the environment, overlooking the scale at which they have participated in all kinds of environmental processes. For example, the Everglades restoration project in Florida is considered as an exemplar for adaptive management[①] framework and the best practice in environmental engineering. Established in 1949, the South Florida Water Management District is a regional governmental agency that manages the water resources in the southern half of Florida, including the Everglades, a 1.5 million-acre wetland sustaining numerous wildlife species.[47] This is the ultimate "wilderness" in the eyes of urban residents. However, the wild Everglades is, in fact, a highly maintained place. In South Florida, numerous sensing stations are installed across the water bodies generating real-time hydrological and water quality data used to build simulation models of the water system. The South Florida Water Management Model is one of the models to analyze operational changes to the water system and to inform management decision making. Moreover, thousands of miles of engineered canals and pipes are carved into Florida's landscape, and water control infrastructures such as water basins, spillways, weir gates, pumps, dams,

---

① 适应性管理可以理解为一种边做边学的决策方法，需通过严格的监控和执行来实现。

① Adaptive management can be understood as a decision-making method via learning by doing, which is achieved by intensely monitoring and actuating processes.

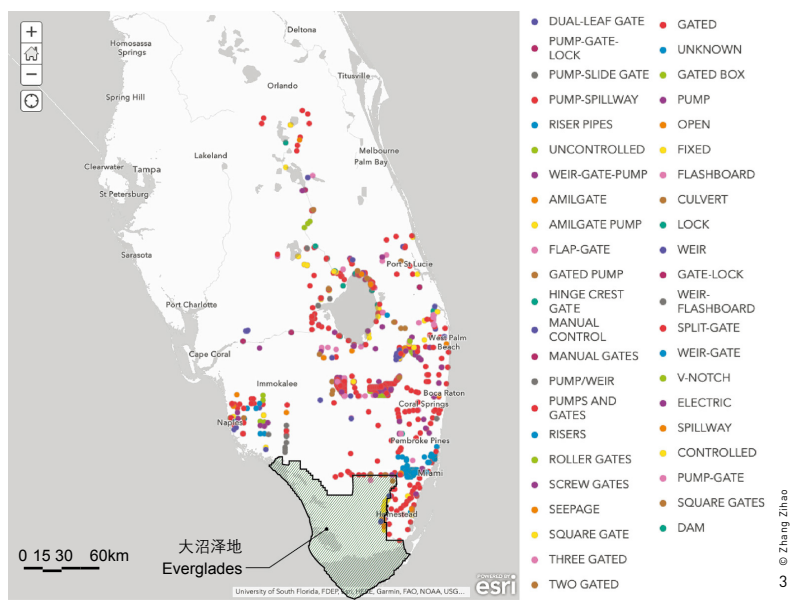

3. 南佛罗里达州水域管理系统（数据来源：Arc Hydro增强数据、在线ArcGIS数据）

3. South Florida Water Management System (Data source: Arc Hydro Enhanced Database and ArcGIS Online)

他利用哈佛大学设计研究生院响应环境与人工制品实验室的水文地貌模拟沙床测试了"感应—处理—制动"响应框架。沙床中沉积物和水流的输入由4个进料器和一个水泵进行控制，此外还配备了传感和监控设备来收集实时数据，包括工作台上方的微软体感设备Kinect和下游的超声波传感器。而后通过Grasshopper软件中的自定义插件将实时数据输入三维建模软件Rhinoceros 3D中。在一次设计实验中，埃斯特拉达提出了名为"调制器"的执行系统，它由一组连接在伺服电机上的亚克力管组成。每个亚克力管均由伺服电机单独驱动，亚克力管的底部插在沉积物中。转动的伺服电机驱动亚克力管上下移动，进而影响水流形态，在工作台的下游形成不同的地形。再通过传感器实时跟踪地形，并使用高程数据建立数字高程模型，便可在设计软件中识别出一系列高地和低地。上述信息可以为接下来的操作提供依据，例如要在高地上建造更多的土地就需要沉积更多的沙子，要侵蚀高地就需要引入更多的水。[50]埃斯特拉达表示，机器人系统展现出了超越人类水平的实时更新和反馈水平。[51]

埃斯特拉达的实验测试了基于"感应—处理—制动"反馈回路的响应式景观框架的可行性，以期利用环境中的传感器和执行器来提供实时景观策略。而遍布环境中的传感器和执行器也为机器学习算法的

and locks are strategically placed along the waterways. These actuators in the system can directly influence the hydrological patterns in South Florida (Fig. 3). In a way, the amount of water and the hydrological patterns are carefully calculated and controlled using simulation models based on real-time hydrological data, weather forecast, and climate modeling.

The Everglades restoration project is only one instance of numerous efforts to use cyberphysical systems to manage environmental processes. The recent discourse of smart cities has fueled cybernetic imaginaries across social sectors. Like Norfolk in the above example, many cities are retrofitting their stormwater systems with controllable and smart actuators provided by emerging companies such as Opti and Xylem that are specialized in live environmental monitoring and real-time control solutions.[48][49] To some extent, the environment is laden with intelligent machines to be mobilized by designers. Two speculative landscape design research projects and a range of robotic and ML examples shed light on an adaptive framework based on real-time feedback and ML, offering insights for communicating and designing with intelligent machines.

In the first project, Towards Sentience, designer Leif Estrada proposed distributing intelligent machines and sensing networks in the Los Angeles River to influence hydrological patterns and build landforms over time. He tested the sensing–processing–actuating responsive framework on a hydromorphology table at the Responsive Environments and Artifacts Lab in the Graduate School of Design, Harvard University. The sediments and water flow inputs were controlled with 4 material feeders and a water pump. The table was also equipped with sensing and monitoring devices to gather real-time data, including a Microsoft Kinect above the table and ultrasonic sensors downstream. Real-time data then fed into Rhinoceros 3D with Grasshopper plugins and customized interfaces. In one of the design experiments, Estrada proposed an actuating system called "attuner" that consists of a matrix of acrylic dowels connected to servo motors. Every dowel was separately driven by a servo motor, and the bottom portion of the dowel sticks into the sediments. When the servo motors turned on, they drove the dowels moving up and down to change the flow pattern, thus creating different landforms downstream of the table. The topography was then live tracked by sensors, and elevation data were used to build a digital elevation model so that a series of high grounds and low grounds can be identified. This information could inform operations such as building more land in a high ground by depositing more sand, and eroding the high ground away by directing more water.[50] Estrada reported that the cyborg system exhibits a level of live updates and feedback beyond human capacity.[51]

Estrada's experiment tested sensing–processing–actuating feedback loop as a viable responsive landscape framework to utilize sensors and actuators in the environment for deploying real-time landscape strategies. Moreover, the environment laden with sensors and actuators sets the basis for applying ML algorithms. Designers can speculate a DRL agent that can test different policies with the "attuners" and adjust strategies according to real-time feedback.

应用奠定了基础。例如，设计师可以构想一种DRL智能体，利用"调制器"来测试不同的景观策略，并根据实时反馈进行调整；或将不同的DRL智能体嵌入环境中，随时间的推移与其他人类及非人类个体和系统共同演化。

要实现以上构想，需要解决两个问题。首先，只有在"持续学习"策略的支持下，才能促使智能机器与其他人类和非人类行为者在环境中共同演化。然而，大多数AI算法其实是"离线"的，这意味着在完成训练后，上传到网络物理系统中的是一个已经经过调试的模型（如深度学习中的神经网络）。训练过程实质上是通过将神经网络暴露在大量数据中来调整单个神经元的权重。训练完成后，神经网络将被校准到最佳状态，仅根据输入信息进行输出，而不会在运行时学习任何新知识。一旦应用到气候变化等动态环境中，这些智能体将会很快过时。因此，开发持续学习框架是将智能机器应用于环境治理过程中的一大挑战。其次，机器学习分为训练和测试环节。在实际应用智能体之前，科学家需要提供训练数据供行机器学习。在前文所述的DRL案例中，智能体能够在模拟环境（如电子游戏）中寻找策略，而这种环境是可以无限次重启的。但真实环境没有"重启键"，如果某个智能体无故造成了恶劣结果，那么将难以挽回。这也是利用人工智能来管理环境所不得不面对的伦理挑战。

另一个探索性研究项目和人工智能研究的最新进展或许可为这些困境提供新思路。来自麻省理工学院建筑学院的设计师里卡多·詹尼·冈萨雷斯在他的论文中提出了一种嵌于冰冻圈环境中的系统。该系统由一个中央"大脑"（可以理解为超级计算机单元）和分布各处的"身体"（改变物理环境的执行器）组成。在具体操作中，"大脑"首先对"身体"施加不同的执行策略，而后根据预测结果与实际差异来评估策略；再将更新后的新策略投放到"身体"中。产生最佳结果的策略将作为成功经验嵌入下一轮干预中。在这样的模式下，某一"身体"可以通过将知识转移给其他"身体"来彼此影响。经过不断迭代，该分布式系统可以逐渐适应冰冻圈环境并随环境一起演化。[52]

该项目的理论框架反映了一些被广泛使用的机器学习算法和技术，例如持续学习、遗传算法和迁移学习。值得注意的是，持续学习一直都是DRL理论的基础（常与DRL相类比的概念是人类的试错学习，而人类的学习过程需要终身且持续的努力）。实际上，有关人工智能的研究已经在探索如何使智能体边学习边衍变的策略，科学家们也在克服"灾难性遗忘"方面取得了可喜进展。"灾难性遗忘"是持续学习的主要障碍——智能体往往会在获取新知识后突然完全"忘记"之前学习过的信息。[53]在目前的人工智能应用中，理论层面的持

Designers can also speculate various DRL agents that are deeply embedded in the environment and co-evolve with other human and nonhuman agents and systems over time.

This kind of speculation raises two concerns. First, designers need continuous learning strategies to anticipate an intelligent machine that co-evolves with other human and nonhuman agents in the environment. Most of the AI algorithms are considered "offline," which means that after training, what is uploaded to a cyberphysical system is a fine-tuned model—a neural network in the context of deep learning. The training process essentially tunes the weights in individual neurons by exposing a neural network with large amounts of data. The neural network is calibrated after training, and simply generates outputs without learning anything new when in operation. In terms of climate change, an agent will be soon out of date. Therefore, developing a continuous learning framework is the next challenge for incorporating intelligent machines in environmental management. Second, ML consists of training and testing sessions. Scientists need training data for machines to learn from before the agent can be applied in real-world situations. In the DRL case, an agent develops its strategies in a simulated environment (e.g. video game) which can be restarted for unlimited times. However, there is no "reset button" for the environment, and if an agent somehow causes an unwanted result, there is no way to go back. This also poses ethical challenges to applying AI agents in environmental management.

Another speculative research project and some recent advances in AI research provide insights into these predicaments. In an MIT architectural thesis research, designer Ricardo Jnani Gonzalez proposed a system deeply embedded in the cryosphere environment. The system consists of a central "mind," which can be understood as a supercomputer unit, and distributed "bodies," which are actuators that change the physical environment. In operation, the "mind" first casts a vast array of varied actuating policies across the "bodies." Then the "mind" evaluates the policies based on the discrepancies between projected scenarios and actual outcomes. The "mind" then updates and casts the policies to the "bodies" again. If one policy yields the best outcome, the successful experience would be embedded within the next iteration of intervention. In this way, one "body" could influence other "bodies" by knowledge exchange, through which this distributed system can gradually attune to the cryosphere environment and evolve with it.[52]

This project's framework mirrors some widely used ML algorithms and techniques such as continuous learning, genetic algorithms, and transfer learning. It should be recognized that continuous learning has always been a theoretical premise for deep reinforcement learning—DRL is often compared with human learning through trial and error which requires lifelong and continuous efforts. In fact, there are already efforts in the AI community to develop strategies that allow agents to learn and evolve. Scientists are making promising progress in overcoming "catastrophic forgetting," which is the major obstacle to continuous learning—an agent tends to completely and abruptly "forget" previously learned information after acquiring new knowledge.[53] In current AI applications, this theoretical

续学习主要通过不断提高人工智能系统的更新频率来实现。可以预见，在不久的将来，更多的持续学习框架和自动更新机制将会被广泛运用。这就要求设计师相应提出具有足够灵活性和适应性的理论框架以支持这些技术的实践应用。

遗传算法的灵感来源于自然选择过程——将上一代的优秀基因遗传给下一代。与此相类似，在遗传算法的训练中，一系列随机策略中的最佳策略会被保留；在下一代的训练中，这些成功策略经过变异，更新为新一套策略进行应用。随着"选择—变异"过程的循环往复，算法的性能也不断提高。同样，在环境管理方面，设计师不应追求模型的复杂性，以期生成一劳永逸的策略；相反，可以尝试利用可单独实施渐进式干预的分布式智能体，使其共享知识、共同演进。当前已有众多相关研究案例：一款名为"Romu"的机器人可将联锁板桩打入地下来建造拦水坝，在防止水土流失的同时促进干旱地区的地下水补给[54]；另一款名为"RangerBot"的水下视觉式机器人可用于保护大堡礁，它可识别并杀死以珊瑚为食的海星，并监测珊瑚礁的健康状况和水质情况。这类机器人主要依赖机器学习和计算机视觉来识别水下环境中的有害海星[55]。上述机器人都是基于机器学习不断进化并发展策略的分布式"身体"。

冈萨雷斯所设想的系统可将某一"身体"的成功策略转移给其他"身体"，这与人工智能研究中的迁移学习异曲同工，即将从某个领域中获得的知识应用于另一领域。相关案例[56]-[58]可为设计师提供借鉴：一方面，机器在分散投放到环境中之前就可以预先接受一般性知识的训练，因而可以快速适应不同的条件；另一方面，成功经验可以通过迁移学习推广到不同的环境中。

## 2.3 植物–机器交互

景观设计师如何开发能与植物直接互动的工具以在城市环境中现实育野？他们或可从当前的农业实践和艺术创作两种模型中汲取经验。精准农业使用传感器来监控植物，并利用机械臂在单株植物尺度施肥、浇水和收获农产品。许多现代农场已将这些机器纳入日常作业中。随着传感技术和机器人技术的不断发展，诸如FarmBot等初创公司开始专注于构建小型自动化园艺系统。[59]精准农业的目标是通过控制植物生长来提高生产力。但这种单一变量系统仅优化了植物作为生命体的其中一个方面。在某种程度上，精准农业通过消除植物的"野性"来提高生产率，强调"植物为人所用"。

continuous learning process is substituted by updating the AI models with higher frequencies. It should be anticipated that more continuous learning architectures and automated updating mechanisms will be widely and shortly available. This requires designers to develop equivalent conceptual frameworks that are flexible and adaptive enough to incorporate these techniques in practice.

Genetic algorithms are inspired by the natural selection process, in which the best qualities of the parent generation are preserved in the child generation. When training with a genetic algorithm, the parent generation represents an array of random policies, and the most effective ones are preserved. In the next generation, mutations can be introduced to diversify the preserved strategies, and a new array of updated policies can be deployed. Repeating this selection-mutation process can increase the performance of the algorithms. For environmental management, one should not speculate on one complex model that does all the hard labors to compute one-shot policies. Instead, one can imagine many distributed agents making incremental interventions individually, and they can share knowledge and evolve together. There are many examples in related research. Romu, a robot designed to build check dams by driving interlocking sheet piles into the ground, can prevent erosion and promote groundwater recharge in arid regions.[54] RangerBot is a vision-based underwater robot that can identify and kill coral-eating starfish and monitor reef health and water quality, protecting the Great Barrier Reef. It relies on ML and computer vision to identify the unwanted starfish underwater.[55] These robots can act as distributed "bodies" for ML agents to evolve and develop strategies.

Gonzalez's speculation of one body transferring successful policies to other bodies requires a technique in AI research known as transfer learning—a machine learning framework that can transfer knowledge acquired from one domain to another. Related examples[56]-[58] are inspirational for designers: on the one hand, machines can be pre-trained with general knowledge about the system before being distributed in the environment. They can quickly attune to different conditions. On the other hand, successful experiences can be generalized and applied to another case.

## 2.3 Plant–Machine Interaction

How can landscape architects cultivate wildness by developing tools that directly interact with the plants in urban environments? There are two existing models: one in agriculture and the other in art. Precision farming uses sensors to monitor the plants and robotic arms to fertilize, water, and harvest at an individual plant level. Many modern farms have started to incorporate these machines into daily practice. With advances in sensing and robotic technologies, startup firms focus on small-scale robotic gardening systems, such as FarmBot.[59] The goal of precision farming is to manipulate plants for a greater productivity. It is a single variant system that only optimizes one aspect of plant life. In a way, precision farming increases productivity by eliminating "wildness" in plants; plants are exploited for human use.

Another model is in art. Artist David Bowen used cybernetic technologies to create art projects that explore the relationship between machines and plant materials. Growth Rendering Device (2007) was such a feedback system: A pea plant was suspended in a hydroponic solution in a bottle attached to a vertical scanner, a roll of paper, an inkjet printer head, and a light. The system provided food and light to the plants and recorded their growth with drawings. A drawing was produced every 24 hours, and then the system scrolled the roll of paper and started the next drawing cycle. As Bowen suggested, this work's outcome was not predetermined because it might record growth, decay or demise of the plant.[60] Based on a similar principle, Growth Modeling Device (2009) scanned a plant from three different angles and 3D printed the plant every 24 hours. Instead of scrolling a paper roll, a conveyor belt advanced approximately 17 inches after each printing.[61] In the art project Plant Drone (2019), Bowen made a plant pilot: A drone was mounted with a plant whose leaves were attached to electrodes. Bioelectricity emitted from the plant leaves became the input to the drone's movements. The plant piloted the drone under the night sky, and the movements were captured with long-exposure photography.[62]

Art production is commonly assumed as human endeavors. Nevertheless, in these projects, the drawings, models or photographs were not completely determined by the artist or the machines; instead, they were co-produced results by machines and plants. Bowen hacked machines and built systems that allowed plants to express themselves and exercise their agency to produce art.

These examples show that it is possible to develop cybernetic systems that bypass control and optimization. Algorithmic Cultivation, a prototyping project developed by designers from the University of Virginia[②], is a platform consisting of robotic armatures, lighting systems, and planters (Fig. 4). The robotic armatures can be equipped with sensors and customized actuators to directly interact with

在艺术创作方面，艺术家戴维·鲍温利用智能技术来创造装置艺术，进而探索机器与植物之间的关系。他于2007年建造的"植物生长描绘装置"建立了机器与植物之间的反馈系统。鲍温将豌豆苗置于一瓶水培营养液中，并将瓶子挂在垂直扫描仪上，其他工具还包括一卷纸、喷墨打印机针头和日光灯。该装置为植物提供养分和光照，并通过扫描数据绘制图纸，记录植物生长。装置每24小时完成一次绘图，而后纸卷滚动，开启下一个绘图周期。正如鲍温所述，该艺术装置的绘图结果无法预先确定，因为它可能记录植物的生长，也可能记录植物的死亡和凋零。[60]基于类似原理，鲍温于2009年创作了"生长建模装置"，分别从三个不同的角度扫描植物，再利用3D打印技术每24小时打印植物的形态。新装置不再设纸卷，取而代之的是一条传送带，每次打印完成后传送带前进约45cm，开启下一个周期。[61]在2019年创作的另一个艺术装置"植物无人机"中，植物开始充当飞行员的角色：鲍温将一株植物安装在无人机中，叶片与电极相连，植物叶片发出的生物电控制无人机飞行方向。由植物"驾驶"的无人机在夜空中飞行，长时间曝光摄影技术将飞行轨迹记录了下来。[62]

人们通常认为艺术创作仅限人类所为。然而，在以上这些艺术装置中，最终生成的绘作、模型及照片并非完全由艺术家或机器决定，它们是机器和植物共同的作品。鲍温通过破解机器原理来制造全新的系统，让植物发挥能动性，实现植物的自我表达和艺术创作。

上述案例表明，智能系统的开发可以绕开"控制"和"优化"这两个概念。来自弗吉尼亚大学的设计师开发了一个名为"算法种植平台"的原型设计项目[②]，平台由机械装置、照明系统和种植槽组成（图4）。机械装置可配备传感器和自定义执行器来直接与植物互动（如修剪和浇水）。该项目的目标是构建机器与植物之间的松散耦合，而非利用机器来促进植物生长。研究团队期望观察植物如何对机器的算法管理做出反馈，而植物的反馈又如何推动机器适应和调整其

② "算法种植平台"由耿百利、罗宾·德里普斯、露西亚·菲尼和张雯菁开发，该项目是弗吉尼亚大学人文信息实验室资助的智慧环境小组所开发的5个项目之一。

② Algorithmic Cultivation was developed by Bradley Cantrell, Robin Dripps, Lucia Phinney, and Emma Mendel. It is one of the 5 projects developed within the Smart Environment Group, which is funded by Humanities Informatics Lab, University of Virginia.

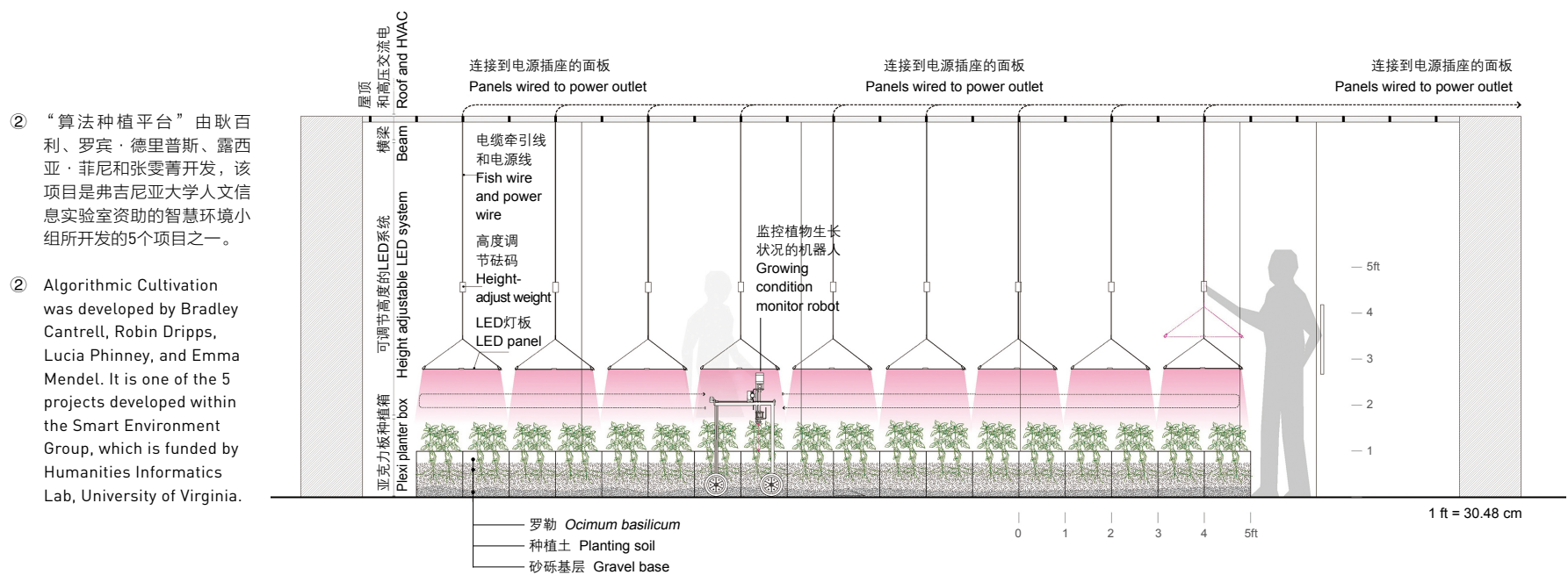

4-1

策略。机器和植物之间将形成正反馈回路，共同演生出离轨的行为。由此产生的结果可视为"高度培育的野"，因为其间的管理逻辑并非人类所熟知或易于理解的（图5）。

从概念上讲，算法种植平台介于工程手段（如精准农业）与艺术表达（如鲍文的艺术装置）之间。一方面，该项目并非可产品化的工业原型。当前市场上已有众多类似系统，尤其是在精准农业和机器人种植领域。另一方面，与概念化的艺术表达不同，算法种植平台是一个持续的探索过程，具有实用价值。该项目为师生们提供了一个聚焦于机器和植物之间相互作用的景观策略测试平台，它也值得设计师更进一步在艺术与科学间进行探索。我们可以将这一项目中应用的方法理解为"后原型开发"，它利用一种全新的框架来解读人们所熟知的主流实践，使其变得陌生，进而鼓励人们重新理解这些实践并探寻更多未知的领域。

## 3 结语：迈向技术多样性与机器的"野"

本文认识到荒野概念的多元性，鼓励设计师通过构建"新形式的野"来拓展对自然和荒野的认知。设计城市荒野并非要建造看起来"自然"和"野生"的风景或恢复历史生态格局。本文认为，应当探索不同于仅利用自然过程来实施再野化的育野策略：通过促进各种非人物种和智能机器的自主性，让它们也成为创造景观的积极协作者。基于该理念，本文探讨了智能机器如何能够大力推进城市环境中野地的构建，并基于科学、工程和景观研究实例，构想了在景观和环境实践中探讨机器智能的理论框架。

该框架的核心——"技术多样性"和"机器生态"也是哲学家许煜的关键思想，可以帮助我们反观本文所述案例。许煜认为，物种的多样性是生物间生态关系的基础，因而生态取决于生物多样性[63]。相

4-1. 算法种植平台装备示意图
4-2. 算法种植平台装备实景图

4-1. Schematic diagram of the Algorithmic Cultivation platform setup
4-2. Photo of the Algorithmic Cultivation platform setup

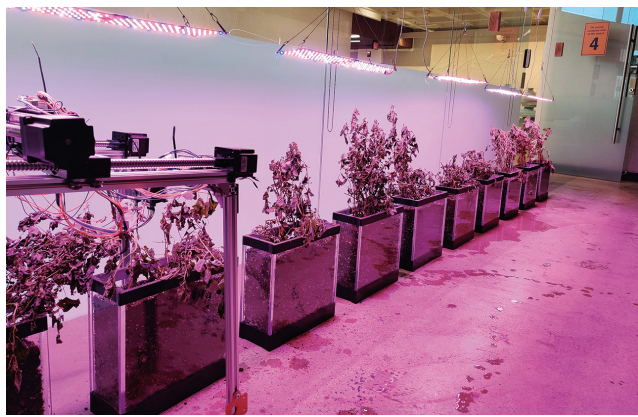

4-2

the plants, such as pruning and watering. Rather than using machines to optimize plant growth, this project aims to develop loose couplings between machines and plants. The team is interested in observing how plants are responding to algorithmic management and, in turn, how the machine starts to adapt and attune to the plants' growth. The machine and plants would form positive feedback loops that, together, produce spin-off and emergent behaviors. The outcome can be said as highly cultivated wildness, as the logic for management is foreign and not easily understood by human observers (Fig. 5).

Conceptually, Algorithmic Cultivation sits between engineering approaches such as the precision farming and artistic expressions such as Bowen's installations. On the one hand, the project is not an industrial prototype that can be monetized as a product. There are already a plenty of similar systems in the market, especially when considering precision farming and robotic agriculture. On the other hand, unlike artistic expressions that function at a conceptual level, Algorithmic Cultivation is an ongoing process with a sense of pragmatism. The project establishes a platform for faculties and students to test out landscape strategies that focus on the interactions between machines and plants. From this vantage, it merits further explorations between art and science. The technique used in this project can be conceptualized as *post-prototyping* because it denaturalizes mainstream practices and reframes them in a different discourse to cultivate alternative understandings and reveal unexplored territories.

## 3 Conclusion: Towards Technodiversity and Wildness in Machines

This paper recognizes the plurality in the wilderness concept and challenges designers to expand the conception of nature and wilderness by constructing "new wilds." Designing urban wilds is not about constructing sceneries that look "natural" and "wild" or restoring historical ecological patterns. Instead, this paper proposes an alternative framework for rewilding to what is practiced with natural processes alone. Cultivated wildness suggests a distinct new type of practices which is about promoting the autonomy of different nonhuman species along with machine agents which are all active co-conspirators in shaping the landscapes. With this conviction, the paper explores how intelligent machines can be important actors in constructing wild experiences in urban environments. Drawing examples from science, engineering, and landscape research, the paper maps out a framework to conceptualize machine intelligence in landscape and environmental design practices.

At the heart of this framework is a sense of "technodiversity" and "ecology of machines," which are two key concepts in philosopher Yuk Hui's thinking that can help us reflect on the cases presented in this paper. Hui's notion of ecology is based on biodiversity because the diversity in species is the fundament for ecological relations between beings.[63] Consequently, "ecology of machines" relies on technodiversity. The advance of one technique as a universal solution (e.g. precision farming becoming the go-to approach in agriculture) means the elimination of other cultural techniques: there was technodiversity before precision farming became the

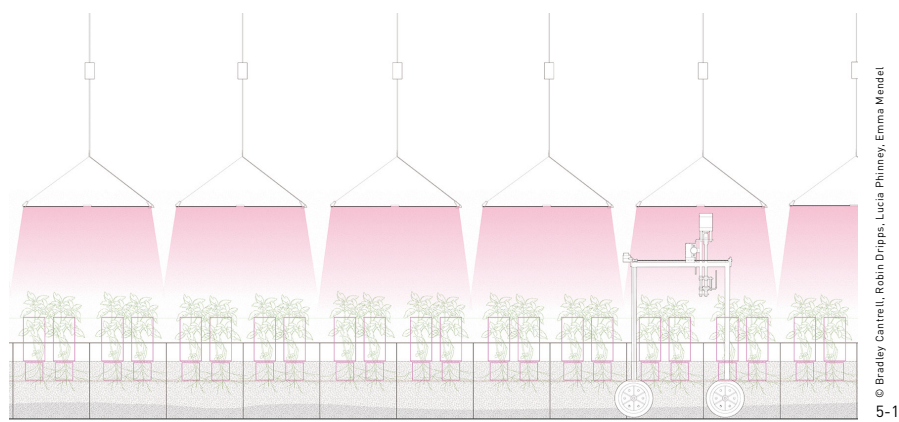

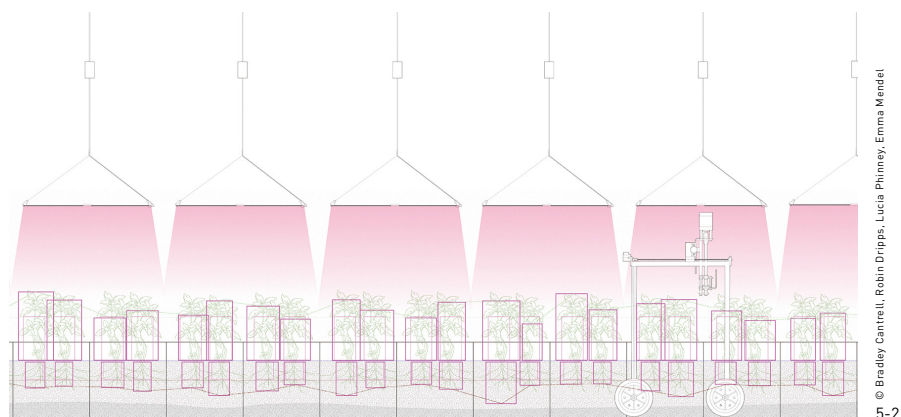

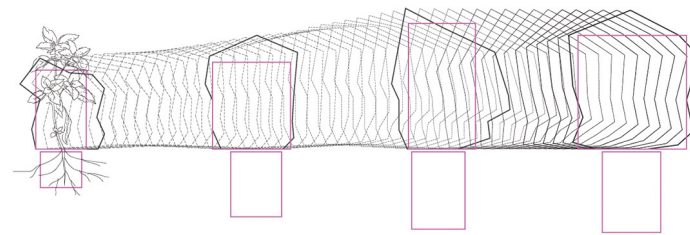

5. 概念图显示，随着时间的推移，不同植物个体对算法培育策略的反馈各有差异，且植物形态和株型也呈现出"野性"（紫色方框表示植物的地上及地下部分）。

5. Conceptual diagrams showing individual plants responding to algorithmic cultivation strategies differently over time and exhibiting wildness in terms of plant form and architecture (purple boxes in the diagram represent plant shoots and roots).

应的，所谓"机器生态"则取决于"技术多样性"。若将某种技术作为普世方法（如将精准农业作为农业实践的首选手段），便会抹杀其他"文化技术"：这意味着技术多样性仅存在于精准农业成为主流之前。在这里，技术多样性指通过不同的认知框架来理解同一个机器，而非同一个机器的不同使用方法。尽管当下机器种类繁多，但主流技术发展仍依赖单一的技术框架，即不断追求效率，扩大人类在环境中的影响力；这也意味着当前尚未形成"机器生态"。因此，本文通过案例分析，期望在主流技术畅想（实现更好的人为控制）之下唤起人们对技术多样性的思考。

在某种程度上，育野也意味着意识到机器的"野"（即不受控制）。技术是人类与其他物种和环境联系的媒介，但技术发展的单一愿景拘囿了人们的思想，剥夺了物种间本有可能形成的相互作用与联系途径。因此，本文认为，荒野的生物多样性保护离不开技术多样性的探索，我们应当通过多元的技术与框架来超越狭隘的愿景。**LAF**

mainstream solution. Technodiversity does not mean to use a machine in different ways, but to explore a machine with different epistemic frameworks. Even though there are different machines, contemporary mainstream technological development is still a monoculture with a single framework for the optimization of efficiency to extend human influence in the environment; nowadays there is no "ecology of machines." The cases presented in this paper are efforts to cultivate a sense of technodiversity within the mainstream technological imagination built around human control.

To some extent, the notion of cultivated wildness also means recognizing the sense of wildness found in machines. Technologies are media through which humans connect with other species and the environment. But a singular vision on technological development promotes a monocultural approach that eliminates other possibilities that species could connect and interact with each other. Thus, one important aspect of preserving biodiversity in wild places is preserving technodiversity—different techniques and frameworks of using the tools beyond a singular vision. **LAF**

**致谢**

感谢审稿人及《景观设计学》编辑部在同行评议过程中给予本文的大力支持。

**ACKNOWLEDGEMENTS**

The authors thank the reviewers' engagement and the support by editors of *Landscape Architecture Frontiers* journal in the exchange of ideas in the peer-review process.